\documentclass[12pt,letterpaper]{IEEEtran}
\usepackage{algorithm,algorithmic}
\usepackage{graphicx,subfigure,booktabs}
\usepackage{flushend}
\usepackage{setspace}

\usepackage{color}
\usepackage{amsmath,amssymb}
\usepackage{colortbl}
\usepackage[bookmarks=false]{hyperref}

\definecolor{HRT}{rgb}{0,0,0}

\title{Fast Barcode Retrieval for Consensus Contouring}

\author{
 \authorblockN{ H.R.Tizhoosh\authorrefmark{1}, G.J.Czarnota\authorrefmark{2}\\ \vspace{0.1in}
}
\authorblockA{\authorrefmark{1} KIMIA Lab, University of Waterloo, Waterloo, ON, Canada, tizhoosh@uwaterloo.ca\\}
\authorblockA{\authorrefmark{2} Department of Radiation Oncology, Sunnybrook Health Sciences Centre\\Toronto, ON, Canada,  Gregory.Czarnota@sunnybrook.ca
}}

\begin{document}
\maketitle
\pagenumbering{gobble}
\begin{abstract}
Marking tumors and organs is a challenging task suffering from both inter- and intra-observer variability. The literature quantifies observer variability by generating consensus among multiple experts when they mark the same image. Automatically building consensus contours to establish quality assurance for image segmentation is presently absent in the clinical practice. As the \emph{big data} becomes more and more available, techniques to access a large number of existing segments of multiple experts becomes possible. Fast algorithms are, hence, required to facilitate the search for similar cases. The present work puts forward a potential framework that tested with small datasets (both synthetic and real images) displays the reliability of finding similar images. In this paper, the idea of content-based barcodes is used to retrieve similar cases in order to build consensus contours in medical image segmentation. This approach may be regarded as an extension of the conventional atlas-based segmentation that generally works with rather small atlases due to required computational expenses. The fast segment-retrieval process via barcodes makes it possible to create and use large atlases, something that directly contributes to the quality of the consensus building. Because the accuracy of experts' contours must be measured, we first used 500 synthetic prostate images with their gold markers and delineations by 20 simulated users. The fast barcode-guided computed consensus delivered an average error of $8\%\!\pm\!5\%$ compared against the gold standard segments. Furthermore, we used magnetic resonance images of prostates from 15 patients delineated by 5 oncologists and selected the best delineations to serve as the gold-standard segments. The proposed barcode atlas achieved a Jaccard overlap of $87\%\!\pm\!9\%$ with the contours of the gold-standard segments. 
\end{abstract}



\section{Introduction}
It is critical to accurately delineate the regions of interest (ROIs) in medical images for computer-assisted analysis. These ROIs identify the targets for therapy or the areas of interest when monitoring therapy. In the former case, delineation is important to ensure that the disease is treated and that normal tissue is spared. In the latter case, delineation is important to ensure that there is accurate feedback from the treatment. When targeting disease for medical procedures in terms of surgical management, it is critical that the boundaries are accurate to the millimeter. Similarly, in radiation oncology, the therapeutic accuracy for targeting the gross tumor is typically within 5 mm for microscopic disease extending from the gross tumor, but within 1--2 mm for high-precision stereotactic radio-surgery. With regard to monitoring similar disease identification accurately and with precision remains important. It is important to properly identify responsive and unresponsive regions, in order to avoid poor sensitivity and specificity owing to classification errors.

Manually contouring (i.e., segmenting) the ROIs in medical images is a tedious and time-consuming task, and it is prone to errors and inconsistencies. Automated segmentation, or auto-contouring, is generally used to assist physicians by performing the same task through customized algorithms. Marking ROIs in medical images has diverse objectives. Whereas in \emph{diagnostic radiology}, one can simply highlight the approximate vicinity containing the ROI (e.g., by drawing an arrow pointing to the ROI, or by inserting some landmarks on the boundary of the ROI), in \emph{treatment planning}, such as for radiotherapy, the exact boundaries of the ROI are required (i.e., by drawing a contour around the ROI in 2D slices). The latter objective---that is, to accurately find the contours around tumors, organs, or lesions---is the subject of this paper. 

\begin{figure}[t]
\begin{center}
\includegraphics[width=0.7\columnwidth]{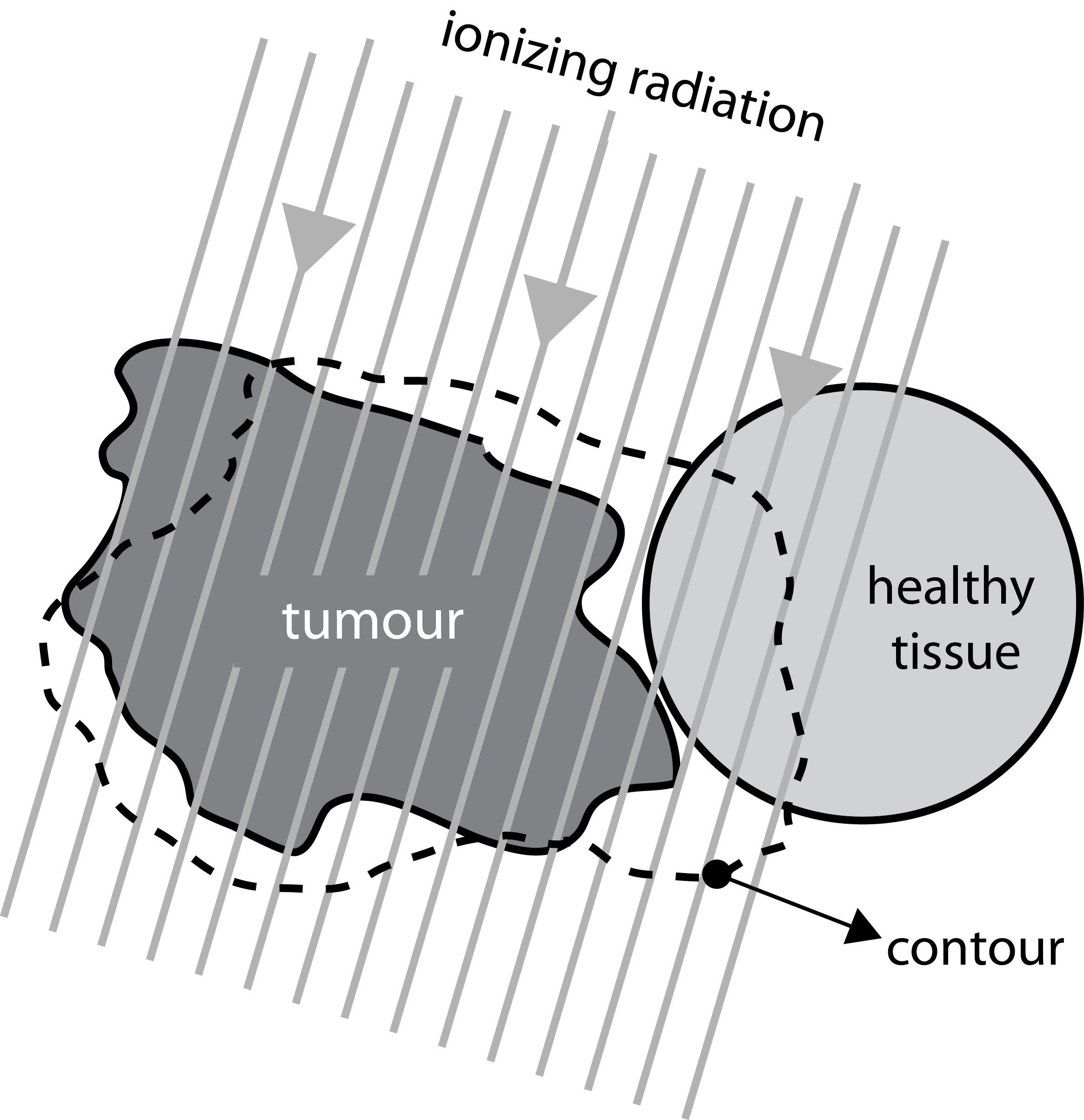}
\caption{The contour (dashed outline) should mark the boundaries of the tumor (irregular shape) that will be treated by radiation. Inaccurate contours prevent efficient treatment, and they can lead to damaged healthy tissue (circle).}
\label{fig:radiation}
\end{center}
\end{figure}

The literature on medical image segmentation is vast and diverse. From simple thresholding steps to sophisticated machine-learning schemes, algorithms have been proposed to segment ROIs in computerized tomography (CT) scans, magnetic resonance (MR) images, and ultrasound images. Owing to the critical nature of medical imaging and the effect of diagnosis and treatment on patients' health, physicians and clinical experts continue to manually contour ROIs in many cases. When automated or semi-automated methods are used, experts always inspect the results, and often edit (i.e., modify) the contour before approving it for use (e.g., for dosimetric calculations in radiotherapy). Hence, the contours used in clinical settings can generally be regarded as the user's output, because they are either created from scratch (with manual delineation), or they are the result of the user editing the results from some algorithm. Inaccuracies in contours (i.e., in the outlines of segments) can have significantly negative effects on the patient's health and constitute a considerable financial burden on any healthcare system, owing to the prolonged treatment and side-effects caused by those inaccuracies. For instance, if contours are used for radiation therapy, inaccuracies will not only decrease the efficiency of the process of destroying the malignancy, but they can also result in damage to healthy adjacent tissue \ref{fig:radiation}. Inaccuracies in the contour are the result of so-called observer variability, which is classified into ``intra''- and ``inter-observer variability''. Intra-observer variability is the inconsistency of the same user (expert) when marking the same images. Inter-observer variability is the difference between different users (experts) when they delineate ROIs in same images. Whenever we use the label ``user'' we mean a physician, or a clinical expert who can understand the image content and delineate ROIs based on his anatomical and clinical knowledge/expertise.     

In this paper, we propose an extension to the conventional atlas-based segmentation by using barcodes, in order to generate consensus contours. Our main contribution is a barcode-guided image-retrieval process that makes it possible to search for similar cases in a large atlas quickly. This facilitates efficient searches of several large atlases (i.e., exploiting big image data), increasing the quality of the computed consensus. As well, we do introduce two new methods for barcode generation. 

The paper is organized as follows: In Section \ref{existlit}, the relevant literature is briefly reviewed. Section \ref{atlaspropos} proposes atlases of barcodes for consensus building via fast barcode-guided segment retrieval. Section \ref{experi} describes our experiments on synthetic images. Section \ref{sec:realimages} discusses another set of experiments with prostate T2-weighted MR images of 15 patients marked by 5 oncologists. Finally, in Section 6, we provide a summary to conclude the paper.


\section{Background}
\label{existlit}
\textbf{Autocontouring --} Automated segmentation (or auto-contouring) is the algorithmic approach to marking ROIs in medical images. Sharma et al. provided a long list of segmentation methods, based on gray-level, texture, and histogram features \cite{Sharma2010}. Other widely used segmentation methods include edge-based segmentation, region-based methods (for merging and splitting), model-based segmentation (with an atlas), active contours \cite{Rahnamayan2005}\cite{Fares2014}, graph-based approaches, registration-oriented segmentation \cite{Khalvati2013}, and machine-learning tools (neural networks, K-means, Fuzzy C-means, etc.). Peng et al. provided a survey of graph-theoretical methods \cite{Peng2013}. Such methods involve modelling segmentation by partitioning a graph into several sub-graphs. They reviewed minimal spanning trees, graph-cut methods, graph-cut methods on Markov random field models, and shortest-path methods. In addition, statistical shape models have been widely used for image segmentation \cite{Heimann2009}. Common statistical shape methods include active shape models, landmark-based shape representations, deformable surfaces, and active appearance models. 

To evaluate the results of image segmentation, the Dice coefficient and the Jaccard index have been used extensively to measure the overlap (agreement) between segments and ground-truth images that are generally provided by the human operators \cite{Crum2006}. Udupa et al. proposed three factors for evaluating segmentation methods: precision (reliability), accuracy (validity), and efficiency (viability). However, they note that ``a surrogate of true segmentation'', namely a gold-standard or ground-truth, is needed for precision \cite{Udupa2006}. 

Using auto-contouring can reduce the variability \cite{Martin2013}. However, researchers have resisted proposing auto-contouring as a solution to observer variability. Through extensive testing in some cases---e.g., with MR brain imaging, and with a large number of segmentation algorithms (up to 20 different methods, in one case)---we know that no single algorithm can solve the variability problem by delivering sufficiently good results \cite{Menze2014}. Clinical experts at the end of the processing chain remain in charge of modifying the result, and they will continue to remain so. Hence, observer variability continues to perpetuate, in spite of algorithmic progress in segmentation. 

\textbf{Observer Variability --} The literature on observer variability in medical imaging is not only vast, it is specialized: researchers generally tend to investigate only a single image modality for a specific region of the body. In the example as our case study in this work, the volume of the \emph{prostate gland} is delineated for various reasons. Transrectal ultrasound imaging is a very common modality for this purpose. It has been reported that the prostate volume can vary when patients have large prostates, and in the respective measurements by experts with different levels of experience \cite{Choi2008}. Using images from more than 100 patients, a coefficient of variation (CV) of almost $5.6\pm7$ was measured among experienced users, whereas for a group of both experienced and inexperienced observers, it was more than double this, at CV$=\!12.2\pm7$.  Furthermore, Sandhu et al. reported prostate volume measurements via contouring by a radiation oncologist and five observers on H-mode transrectal ultrasound (TRUS) images with mean deviations of between 4\% and 29\% \cite{Sandhu2012}. Contouring the prostate gland on cone-beam CT images for image-guided radiotherapy, White et al. reported that ``expert observers had difficulty agreeing upon the location of the prostate peri-prostatic interface on the images provided'' \cite{White2009}. They also reported that the mean difference between a CT-derived volume and the mean cone-beam-derived volume was 16\% (range 0--23.7\%). 

\textbf{Using Atlases --} Atlas-based segmentation (ABS) is a widely used approach to medical image segmentation \cite{Cabezas2011} \cite{Phellan2014}. ABS stores images with their ground-truth segmentations, where the latter are generally manual delineations. A query image is registered and then compared with all images in the atlas. The best match, registered to the query, is delivered as the segmentation. Variations of ABS have become popular medical image segmentation methods, but they suffer from several drawbacks. For instance, multiple atlases are generally needed \cite{Wang2012}\cite{Asman2013}. For them to be sufficiently accurate, these atlases must be large, incorporating many variations of the image class in the hopes that one of them is decidedly similar to the query image. This, if possible at all, poses another challenge: as the number and size of the atlases increase, the computational complexity becomes unfeasible in clinical practice. This drawback is particularly pronounced when using non-rigid transformations, which are generally required in order to capture soft tissue deformations in medical images. However, \emph{binary atlases} have been proposed to overcome such problems \cite{PatentQB}. Searching for binary thumbnails in an atlas is an exceedingly faster operation.  

\textbf{Consensus Building--} To measure the variability, a baseline is needed---viz., a consensus contour. As well, in order to build a consensus using the atlas approach, one needs methods to build a consensus contour. The simultaneous truth and performance level estimation (STAPLE) \cite{Warfield2004} is one of the most commonly used algorithms for generating a consensus contour based on multiple user segments. Another recent method in this regard is Selective and Iterative Method for Performance Level Estimation (SIMPLE) to fuse retrieved labels from an atlas \cite{Langerak2010}. In our experiments, we used STAPLE to compute the consensus when we retrieved similar cases. 

\textbf{Hashing-based Image Retrieval --} Searching for images, or image retrieval, is a crucial step in ABS. There are many different approaches to content-based image retrieval  \cite{Smeulders2000}.  Some researchers have proposed using hashing to index images, making retrieval a fast main-memory operation. Among the more promising approaches is locality-sensitive hashing (LSH) \cite{Indyk1998, Gionis1999}. LSH has been used in different forms for scalable image retrieval \cite{Xia2012}. Zhang et al. presented a supervised kernel-hashing technique that compresses a high-dimensional image feature vector into a small number of binary bits with the informative signatures preserved. These binary codes are then indexed into a hash table, such that the images can be retrieved quickly from a large database \cite{Zhang2015}. Weihong et al. proposed a scheme using LSH, and reported results showing that it is possible to scale hundreds of thousands of images \cite{Weihong2009}. Liu et al. proposed a method for representing local features with binary codes \cite{Liu2014}. 

\textbf{Summary of the Literature Review --} We observe that variability is ubiquitous, and that it poses a serious obstacle to quality assurance in image segmentation. Automated segmentation methods, regardless of their performance or how many of them we fuse, do not seem to offer a solution. Indeed, clinical experts continue to overwrite algorithms and edit their results. Nevertheless, atlas-based segmentation appears viable as a way to generate computed consensus contours, provided we have access to multiple atlases, or a large atlas created by multiple users. However, to make computed consensus reliable, we need many examples, and to make it feasible the retrieval must be fast. Moreover, using some binary embeddings along with the hashing approach is promising.   Therefore, in this paper we propose a method for generating consensus by creating and searching atlases of barcodes to retrieve similar cases. Our algorithmic contribution is the ``Atlas of Barcodes,'' applied to a well-known problem of observer variability.   

\section{Methods}
\subsection{Atlas of Barcodes and Segments}
\label{atlaspropos}
From the literature, we borrow the powerful idea of atlas-based segmentation, and especially the notion of binary atlases for efficient image retrieval when operating on an ensemble of multiple atlases. However, our approach is novel in many ways: First, our approach does not use or depend on registration. If the results are sufficiently good without registration, we can always incorporate some non-rigid alignment procedure. Further, we use barcodes, rather than binary images of some sort. There are many advantages to constructing atlases comprised of barcodes of images and their segments, as we demonstrate. Furthermore, we introduce two new approaches to extracting barcodes. Finally, we test our proposed approach using both synthetic and real prostate images. 


\subsection{Barcodes as Image Annotation} 
Barcodes have been used extensively to label products. However, using barcodes for image-processing tasks, and especially for annotating (or tagging) medical images, is a novel idea \cite{Tizhoosh2015}. We consciously use ``annotation'' here although we mostly understand annotations to be textual informations or some sort of markups. The potential benefit from using barcodes for image annotation (if they are as effective as invariant features), is due to their low storage requirements and high processing speed. Even when compared to binary images (i.e., 2D barcodes), processing barcodes can be performed more efficiently. 

\textbf{Radon Barcodes --} Examining an image $f(x,y)\in [0,L-1]$ encoded with $n$ bits ($L\!=\!2^n$) and consisting of some ROIs that are not entirely black ($f(x,y)\neq 0$), the Radon transform $\mathcal{R}_{f(x,y)}\!: \mathbb{R} \times [0,2\pi)\rightarrow \mathbb{R}_+$ for the image can be given as
\begin{equation}
\mathcal{R}_{f(x,y)}\! =\! \int\limits_{-\infty}^{+\infty}\!f(\rho \cos\theta - \varepsilon\sin\theta, \rho \sin\theta + \varepsilon\cos\theta) d\varepsilon,
\end{equation}
where
\begin{equation}
\begin{bmatrix}
\rho \\
\varepsilon
\end{bmatrix} =
\begin{bmatrix}
\cos\theta & \sin\theta\\
-\sin\theta & \cos\theta
\end{bmatrix}
\begin{bmatrix}
x \\
y
\end{bmatrix}.
\end{equation}

\begin{figure}[htb]
\begin{center}
\includegraphics[width=\columnwidth]{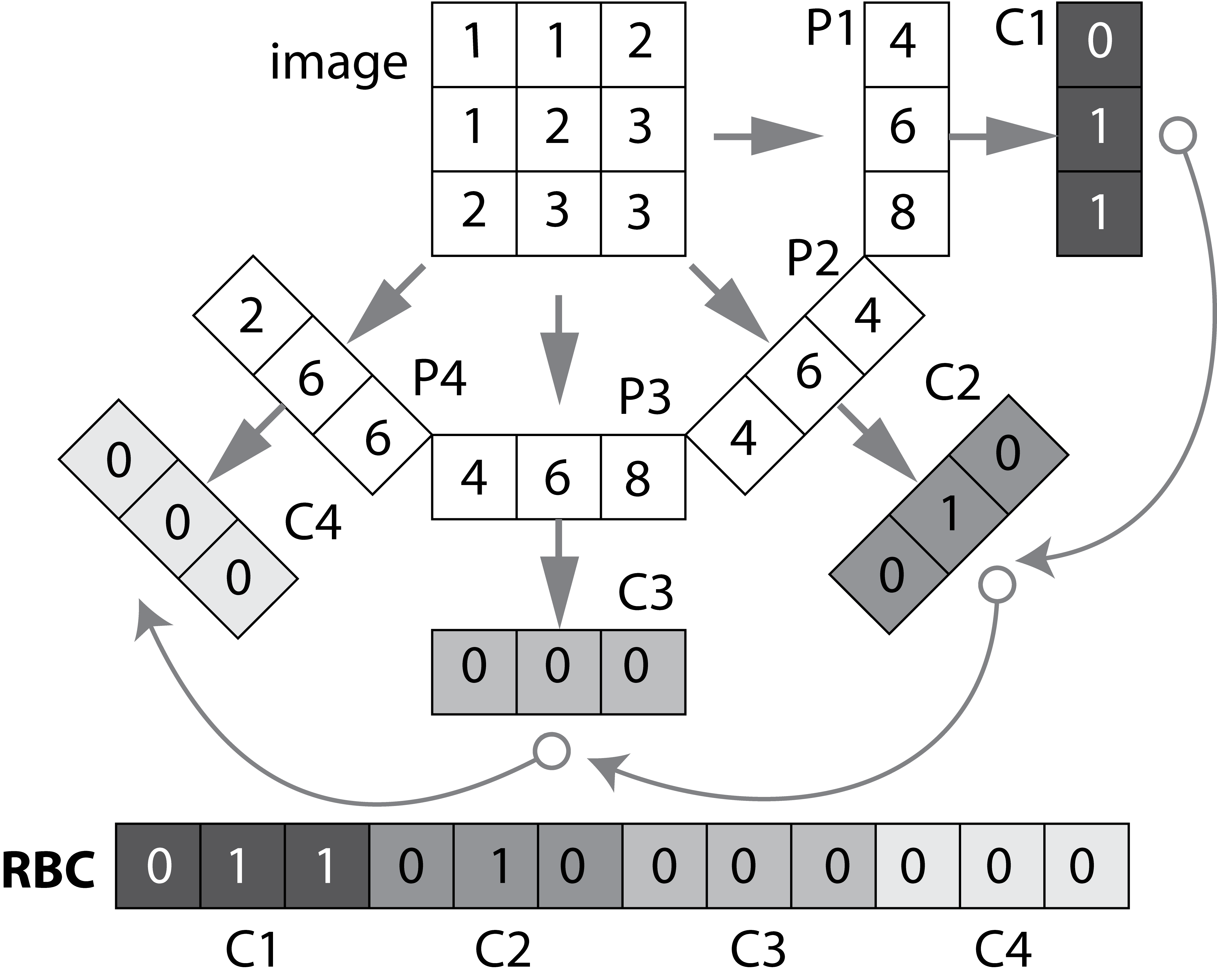}
\caption{Radon Barcode (RBC) -- The image is Radon-transformed. All projections (four, in this example: P1, P2, P3, and P4) are thresholded (via the incremental approach $\mathcal{T}_{\theta=\theta_j}^{\text{incr}} $) to generate the respective code fragments C1, C2, C3, and C4. The concatenation of all code fragments delivers the barcode RBC. As described in Algorithm \ref{alg:Radon}, the thresholding function $\mathcal{T}$ can be implemented in different ways. [Modified graphic similar to \cite{Tizhoosh2015}]}
\label{fig:RBC}
\end{center}
\end{figure}

Tizhoosh proposed generating barcodes by binarizing individual projections \cite{Tizhoosh2015}. For instance, if we threshold all projections for individual angles based on a ``local'' threshold for that angle, then we can assemble a barcode for all thresholded projections, as depicted in Figure \ref{fig:RBC} \cite{Tizhoosh2015}. A simple way to threshold the projections is to calculate a typical value via the median operator applied on all non-zero values of a projection: 
\begin{equation}
\mathcal{T}_{\theta=\theta_j}^{\text{local}} \left(\mathcal{R}_{f(x,y)}\right)=
\begin{cases}
    0,		& \text{if } x\leq \underset{\theta=\theta_j}{\text{median}}\left(\mathcal{R}_{f(x,y)}\right)\\
    1,          & \text{otherwise}
\end{cases}
\end{equation}

Algorithm \ref{alg:Radon} describes how Radon barcodes (RBC) are generated. However, the thresholding function $\mathcal{T}$ can be implemented in various ways. For instance, an ``incremental'' scheme (lines 12--18 in Algorithm \ref{alg:Radon}) binarizes the projections based on the increase of integral values in neighboring lines. With $g(\rho,\theta)=\mathcal{R}_{f(x,y)}$, we get
\begin{equation}
\mathcal{T}_{\theta=\theta_j}^{\text{incr}}  \left(\mathcal{R}_{f(x,y)}\right)=
\begin{cases}
    0,		& \text{if } g(\rho_i,\theta_j) \leq g(\rho_{i+1},\theta_j) \\
    1,          & \text{otherwise}
\end{cases}
\end{equation}

\begin{algorithm}[htb]
\caption{Radon Barcode (RBC) Generation}
\begin{algorithmic}[1]
\label{alg:Radon}
\STATE Initialize Radon barcode $\mathbf{b} \leftarrow \emptyset$ 
\STATE Initialize angle $\theta \leftarrow 0$ and $R_N=C_N=32$
\STATE Normalize image $\bar{I} = \textrm{Normalize}(I,R_N,C_N)$ 
\STATE Set the number of projection angles $n_p$
\STATE Set a thresholding function $\mathcal{T}=$\{local,incremental,global\}
\WHILE{$\theta < 180$}
	\STATE Derive all projections $\mathbf{p}$ for $\theta$
	\IF{local thresholding} 
		\STATE $T_\textrm{typical}\leftarrow\textrm{median}_i (\mathbf{p}_i)|_{\mathbf{p}_i \neq 0}$
		\STATE Binarize projections: $\mathbf{v} \leftarrow \mathbf{p} \geq T_\textrm{typical}$ 
        \ELSE
        		\IF{incremental thresholding}
			\STATE $\mathbf{q} \leftarrow$ ZEROS(length($\mathbf{p}$))
			\FOR{$i=2:$length($\mathbf{p}$)}
				\IF{$\mathbf{q}(i)>\mathbf{q}(i-1) $}
					\STATE $\mathbf{q}(i)\leftarrow1$
				\ENDIF
			\ENDFOR 
		\ELSE
			\STATE No thresholding: $\mathbf{v} \leftarrow \mathbf{p}$ 
		\ENDIF
        \ENDIF
	\STATE Append the new row $\mathbf{b} \leftarrow \textrm{append}(\mathbf{b},\mathbf{v} )$ 
	\STATE $\theta \leftarrow \theta + \frac{180}{n_p}$
\ENDWHILE
\IF{global thresholding}
	\STATE $\mathbf{p} = \mathbf{b}$ 
	\STATE $T_\textrm{typical}\leftarrow\textrm{median}_i (\mathbf{p}_i)|_{\mathbf{p}_i \neq 0}$
	\STATE Binarize projections: $\mathbf{b} \leftarrow \mathbf{p} \geq T_\textrm{typical}$
\ENDIF 
\STATE Return  $\mathbf{b}$
 \end{algorithmic}
 \end{algorithm}

Moreover, a ``global'' approach calculates a threshold for all non-zero values after all of the projections are collected (lines 26--30 in Algorithm \ref{alg:Radon}):
\begin{equation}
\mathcal{T}_{\theta=\theta_j}^{\text{global}}  \left(\mathcal{R}_{f(x,y)}\right)=
\begin{cases}
    0,		& \text{if } x\leq \underset{\theta=\theta_0,\dots,\theta_{179}}{\text{median}}\left(\mathcal{R}_{f(x,y)}\right)\\
    1,          & \text{otherwise}
\end{cases}
\end{equation} 

In order to obtain barcodes of the same length, the function \emph{Normalize$(I)$} resizes all of the images to $R_N\!\times\!C_N$ images (i.e., $R_N\!=\!C_N\!=\!2^n,n\!\in\! \mathbb{N}^+$).

\begin{figure}[htb]
\centering    
\subfigure[Lymphoma]{\label{fig:a}\includegraphics[width=40mm,height=40mm]{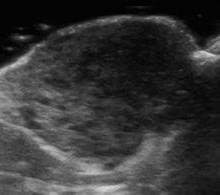}}\qquad
\subfigure[Ganglioglioma]{\label{fig:a}\includegraphics[width=40mm,height=40mm]{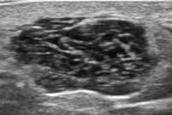}}\\
\subfigure[RBC local]{\label{fig:a}\includegraphics[width=0.45\columnwidth,height=7mm]{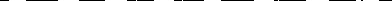}}\qquad
\subfigure[RBC local]{\label{fig:a}\includegraphics[width=0.45\columnwidth,height=7mm]{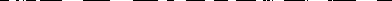}}\\
\subfigure[RBC incremental]{\label{fig:a}\includegraphics[width=0.45\columnwidth,height=7mm]{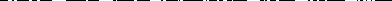}}\qquad
\subfigure[RBC incremental]{\label{fig:a}\includegraphics[width=0.45\columnwidth,height=7mm]{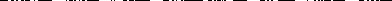}}\\
\subfigure[RBC global]{\label{fig:a}\includegraphics[width=0.45\columnwidth,height=7mm]{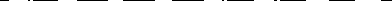}}\qquad
\subfigure[RBC global]{\label{fig:a}\includegraphics[width=0.45\columnwidth,height=7mm]{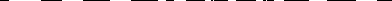}}
\caption{Sample barcodes ($32\!\times\! 32$ normalization). The input image (a) shows a hypervascularized mass that proved to be a B-cell lymphoma. The input image (b) is a benign-looking mass that proved to be a cystic ganglioglioma [Source for the images: http://www.ultrasoundcases.info/]}
\label{fig:Barcodes}
\end{figure}

Figure \ref{fig:Barcodes} depicts barcodes for two breast ultrasound scans.  Calculating the Hamming similarity (see Equation \ref{eq:hamming}) results in a similarity of 76\% for (c)--(d), 75\% for (e)--(f), and 74\% for (g)--(h). 

\subsection{Retrieving Similar Cases via Barcode Matching} 
We built an atlas of barcodes and used the Hamming distance to retrieve a single similar image (only the first hit, namely the most similar case, is considered). The corresponding user segments are also retrieved to compute a consensus contour. No registration is used. Our approach is described in Algorithms \ref{alg:AtlasCreation} and \ref{alg:UsingAtlas}. 

\begin{algorithm}[htb]
\caption{Atlas of barcodes and segments}
\begin{algorithmic}[1]
\label{alg:AtlasCreation}
\STATE /* Initializations */
\STATE Set number of images $n_I$ and user segment/image $n_U$
\STATE Set normalized image size $N\!\times\! N$
\STATE Set barcode ``codeType'', e.g., LBP, RBC 
\STATE /* Create the atlas of barcodes */
\FOR{$i=1$ to $n_I$} 
	\STATE $\mathbf{b}_i \leftarrow$ \emph{calcBarcode($\mathbf{I}_i,N,$ codeType)}
	\STATE Read user segments $\mathbf{S}^1_i,\mathbf{S}^2_i,\dots,\mathbf{S}^{n_U}_i$
	\STATE Save atlas $\mathcal{A}=\{\mathbf{I}_i, \mathbf{b}_i,\mathbf{S}^1_i,\mathbf{S}^2_i,\dots,\mathbf{S}^{n_U}_i\}$
\ENDFOR
\end{algorithmic}
\end{algorithm}

\textbf{Atlas Creation (Algorithm \ref{alg:AtlasCreation}) --} The core of this scheme is found in line 7, during which the function \emph{calcBarcode} calculates a barcode for each image to create the atlas $\mathcal{A}$. Of course, segments $\mathbf{S}_i^1,\mathbf{S}_i^2,\dots,\mathbf{S}_i^{n_U}$ prepared by $n_U$ users (viz., clinical experts and physicians) must be attached to image $\mathbf{I}_i$ and its barcode $\mathbf{b}_i$.  

\textbf{Retrieving Similar Cases (Algorithm \ref{alg:UsingAtlas}) --}  In practice, the clinical user is working on a query image $\mathbf{I}_q$, and has (somehow) generated/edited a segment $\mathbf{S}_q$ (lines 3--4). We first annotate the image with a barcode $\mathbf{b}_q$ (line 6). Then, we begin searching the atlas (lines 7--14), during which the similarity of barcode $\mathbf{b}_q$ for the query image $\mathbf{I}_q$ to each barcode $\mathbf{b}_i$ in the atlas is calculated via the Hamming distance:   
\begin{equation}
h(\mathbf{b}_i,\mathbf{b}_q) = 1 - |\textrm{xor}(\mathbf{b}_i,\mathbf{b}_q)|/|\mathbf{b}_q|.
\label{eq:hamming}
\end{equation}

\begin{algorithm}[htb]
\caption{Find similar cases to build consensus}
\begin{algorithmic}[1]
\label{alg:UsingAtlas}
\STATE Read normalized image size $N\!\times\! N$
\STATE Read barcode ``codeType'', e.g., LBP, RBC 
\STATE Read a new query $\mathbf{I}_q$ from a user
\STATE Read the user's segment $\mathbf{S}_q$
\STATE $\mathbf{b}_q \leftarrow$ \emph{calcBarcode($\mathbf{I}_q,N,$ codeType)}
\STATE $\eta_{\max}=0$
\FOR{$i=1$ to $n_I$}
               \STATE $\mathbf{b}_i \leftarrow$ \emph{readAtlas}($\mathcal{A},i$)
               \STATE$\eta=$ \emph{HammingSimilarity}($\mathbf{b}_i,\mathbf{b}_q$)
                \IF{$\eta > \eta_{\max}$}
                    \STATE $\eta_{\max} = \eta$
                    \STATE \emph{bestMatchIdx} $= i$
                \ENDIF
\ENDFOR
\FOR{$j=1$ to $n_U$}
	\STATE allSegments($j$).$S$ $\leftarrow$ $S^j_\textrm{bestMatchIdx}$
\ENDFOR
\STATE $\mathbf{C} \leftarrow $ \emph{calcConsensus}(allSegments) 
\STATE Display consensus, and report the accuracy $J(\mathbf{S}_q,\mathbf{C})$
\end{algorithmic}
\end{algorithm}

All user segments attached to the best match are then retrieved (lines 15--17).  The consensus segment $\mathbf{C}$ is then computed via the function \emph{calcConsensus}---that is, the STAPLE algorithm \cite{Warfield2004} implemented as described in \cite{Allozi2010}.


\section{Results}
\label{experi}
In the recently published paper that introduced Radon barcodes, the IRMA dataset with 14,400 x-ray images was used for validating the performance content-based barcodes \cite{Tizhoosh2015}. However, these images are of general diagnostic value and have not been segmented. Hence, we cannot use IRMA data, or any other set of images if they have not segmented by multiple experts. Such databases are still a rarity. Databases of segmented medical images do exist (e.g., see segmentation challenges in the MICCAI conferences), but they provide only one ground truth per image. Obviously, as we are proposing to use image retrieval to build the ground-truth (i.e., the consensus), we cannot use such databases. 
  
In what follows, the experiments to validate the proposed atlas of barcodes are described. Our investigation was not concerned with generating consensus for segments with considerable shape irregularities. Indeed, we used the prostate gland as an example because it has a relatively well-defined shape, despite the fact that it is nevertheless subject to considerable observer variability. All of our experiments were performed on an iMac with a 2.93 GHz Intel Core i7 processor and 16 GB 1333 MHz DDR3 memory. 

\subsection{Image Data: Synthetic TRUS Images} 
It is a considerable challenge to validate any approach to computed consensus contouring. The ultimate test, to be sure, is to measure the accuracy of the computed consensus contour $\mathbf{C}$. However, this depends on assessing the quality of the consensus. Unless there is a ``gold-standard segment'' $G$ for each image,  reliable validation will not be straightforward, and any particular observation will be inconclusive. Of course, this is not feasible with real images, for which there is no gold standard. Hence, we generated synthetic images whose gold segments were known a priori. For this reason, we used synthetic images that simulate transrectal ultrasound (TRUS) images.

TRUS images of prostates may be used to both diagnose and treat prostate diseases such as cancer. Starting with a set of prostate shapes $P_1, P_2,\dots, P_m$, we created random segments $G_i$ through combinations of those priors, adding noise along with random translations and rotations, and we distorted the results with speckle noise and shadow patterns. Each image $I_i$ is thus created from its gold $G_i$. Consequently, we can simulate $k$ user delineations $S_i^1, S_i^2,\dots, S_i^k$ by manipulating $G_i$ via scaling, rotation, and morphological changes, and we can simulate edits by running active contours with variable user-simulating parameters. The variability of user delineations was simulated according to several factors: error probability ($[0,0.05]$), anatomical difficulty ($=0.2$ out of $[0, 1]$), and the scaling factor for morphology (form $1\!\times\!1$ to $21\!\times\!21$). The user was modelled according to the level of experience (a random number from $(0,1]$), the user's attention  (a random number from $[0,1]$), and the user's tendencies in terms of the segment size (a random number from $[-1,1]$), whether tending to draw contours that are relatively small ($\rightarrow\!-1$) or large ($\rightarrow\!+1$). 

\begin{figure*}[h]
\begin{center}
\includegraphics[width=1.7in,height=1.7in]{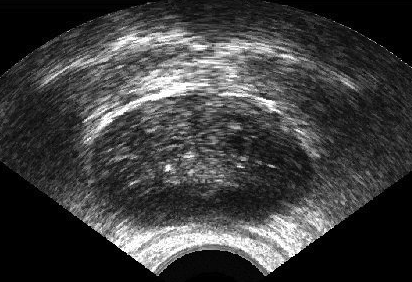}
\includegraphics[width=1.7in,height=1.7in]{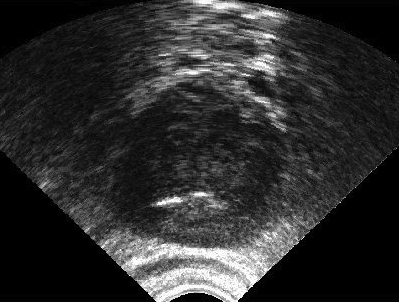}
\includegraphics[width=1.7in,height=1.7in]{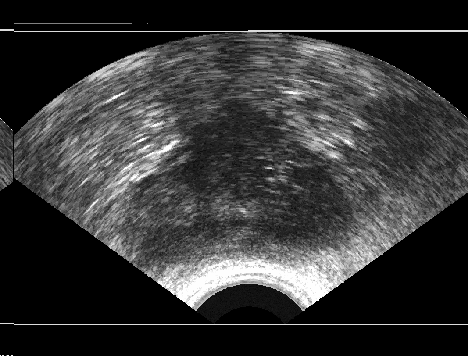} \\ \vspace{0.05in}
\includegraphics[width=1.7in,height=1.7in]{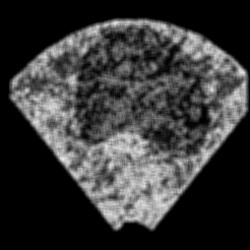}
\includegraphics[width=1.7in,height=1.7in]{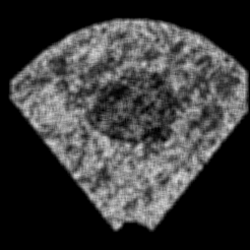} 
\includegraphics[width=1.7in,height=1.7in]{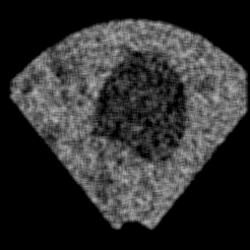} 
\caption{Sample TRUS (top) and simulated images (bottom).}
\label{fig:TRUSsample}
\end{center}
\end{figure*}

\begin{figure*}[h]
\begin{center}
\includegraphics[width=1\textwidth]{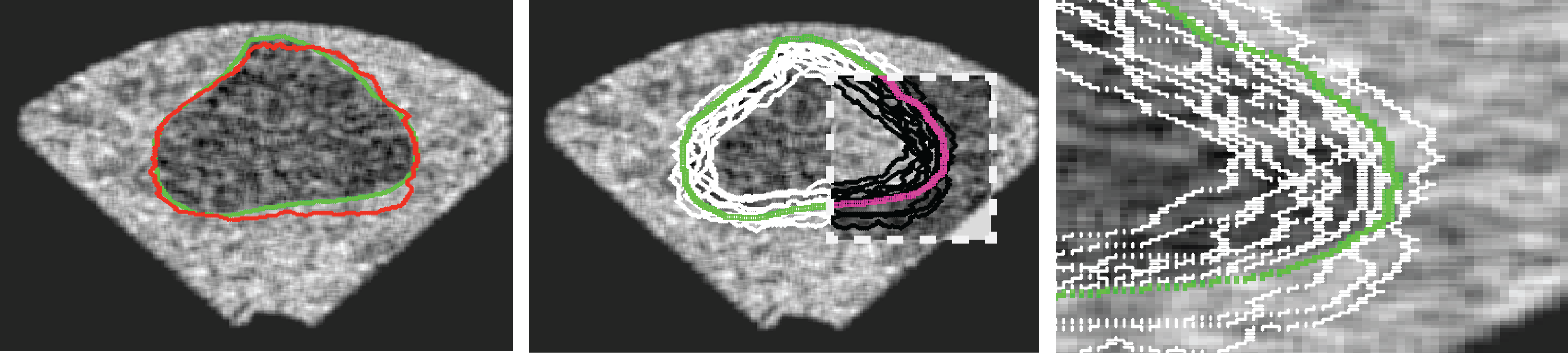}
\caption{Sample image shows gold segments and consensus contours (left). The users have drawn the contour differently (middle, with the gold contour superimposed). The inverted region (middle) is magnified (right) to emphasize details of the variability. \vspace{0.3in}}
\label{fig:sampleImages}
\end{center}
\end{figure*}

We generated 500 images from their corresponding gold-standard images. Furthermore, we generated 20 different segments for each image, assuming that there were 20 users. Figure  \ref{fig:TRUSsample} shows three examples of real and synthetic TRUS images. One should bear in mind that the purpose here was not to simulate the images realistically, but rather to have a base from which to generate variable segments from a perfect segment. Figure \ref{fig:sampleImages} shows an example of the gold segments and simulated user contours. The variability, coupled with the gold segment, is what is needed to validate our approach.    

All images and their user segments use din this paper are publicly available\footnote{\url{http://kimia.uwaterloo.ca}}.

\subsection{LBP versus RBC} 
Local binary patterns (LBPs) \cite{ojala2002} have been used extensively for image classification. LBPs are typically used to calculate a histogram for extracting features. Here, LBPs are extracted and recorded as barcodes through a concatenation of binary vectors around each pixel. Of course, LBPs are not designed to be used as barcodes. Owing to their spatial (window-based) nature, they generate long binary vectors, even when the image is undersampled. We implemented LBPs as barcodes because, as ``binary descriptors,'' they have been remarkably successful.  

Because atlases of images and their barcodes will be generated, it might be asked whether there would be a significant difference if we would use different barcodes to retrieve similar images. Thus, we compared the performance of LBPs and RBCs. Table \ref{tab:LBPRBCconfig} lists the parameter settings. In general, and relatively independent of the number of Radon projections $n_p$, the size of LBP barcodes was larger than RBCs. This, of course, will manifest itself computationally, as the numbers show. 

Comparison with LBP may appear unjustified as LBP was originally design to locally characterize textures. However, countless papers do report the recognition capability of LBP. One has to bear in mind that we apply LBP to find similar images, and not to detect similar segments.  

\begin{table}[b!]
\small
\caption{Settings for using LBPs and RBCs.}
\begin{center}
\begin{tabular}{@{}*{3}{l}@{}}
\toprule
 						& LBP			& RBC	\\
\cmidrule(r){2-3}
Size of normalized image		& $32\times 32$ 	& $32\times 32$ \\
Parameters (window size/rays)	     & $3\times 3$		& $n_p=8$ \\
Number of bits				& $8100$		& $392$ \\ 
\bottomrule
\end{tabular}
\end{center}
\label{tab:LBPRBCconfig}
\end{table}%

To compare barcodes (and other methods), we also calculated the maximum achievable accuracy (MAA), which is the upper bound of Jaccard values for all consensus segments with all gold segments in the atlas. Because we are experimenting with synthetic images, we have the ``prefect'' gold standard segments for each image; if we build the consensus of all user segments for a given image and compare it against this gold standard segment (which is not available for real images), then we can achieve the highest possible accuracy that we can all MAA. Any method that operates based on finding the most similar image and taking the consensus of corresponding $n_U$ user segments without any registration cannot surpass these upper bounds. Of course, calculating the MAA in practice is impossible, because gold segments are only available in experimental settings. 

Table \ref{tab:LBPvsRBC} compares the performance of different barcodes using ``leave-one-out'' validation. As the number of images in the atlas grew, all barcodes converged toward a certain limit. No statistical significance could be detected among all barcodes with respect to their Jaccard index $J$ (i.e., their respective agreement with the gold segment). The times for searching the atlas varied drastically, however, between LBPs and RBCs. The latter was decidedly faster (owing to a shorter bit-string length). RBC-based searching with incremental thresholding appears to be the fastest approach. It is approximately $22$ times faster than LBP-based searching among the 500 images. This might be insignificant when performing a single task. However, for big image data that depend on distributed computing, RBCs are far less computationally expensive.     

\begin{table*}[htb]
\caption{Leave-one-out experiments using barcode-based image retrieval for consensus building among $n_U=20$ users. The accuracy is measured by comparing the computed consensus $\mathbf{C}$ to the gold segment $\mathbf{G}$ via the Jaccard index $J(\mathbf{C},\mathbf{G})$. The maximum achievable accuracy (MAA) for varying the first $n_I$ images in the dataset is provided for comparison. The average search time $\bar{t}$ is in milliseconds. }
\begin{center}
\begin{tabular}{@{}*{10}{c}@{}}
\toprule
 & & \multicolumn{2}{c}{RBC Local} & \multicolumn{2}{c}{RBC Incremental} & \multicolumn{2}{c}{RBC Global} & \multicolumn{2}{c}{LBP}\\
 \cmidrule(r){3-4}\cmidrule(r){5-8}\cmidrule(r){9-10}
 
$n_I/n_U$	 &  \cellcolor[gray]{0.8} MAA & J & $\bar{t}$ & J  & $\bar{t}$ & J  & $\bar{t}$ & J  & $\bar{t}$ \\ 
\cmidrule(r){2-2}\cmidrule(r){3-4}\cmidrule(r){5-6} \cmidrule(r){7-8}\cmidrule(r){9-10}
10/20	& \cellcolor[gray]{0.8} 78$\pm$6 & 75$\pm$6 & $<$1 & 76$\pm$6 & 3$\pm$5 & 75$\pm$6 & 3$\pm$4 & 77$\pm$7 & 4$\pm$19 \\
20/20	& \cellcolor[gray]{0.8} 80$\pm$6 & 78$\pm$7 & $<$1 & 77$\pm$7 & 3$\pm$5 & 76$\pm$6 & 3$\pm$4 & 80$\pm$7 & 4$\pm$19 \\
50/20	& \cellcolor[gray]{0.8} 84$\pm$5 & 79$\pm$8 & 1$\pm$3 & 77$\pm$7 & 3$\pm$5 & 78$\pm$8 & 3$\pm$4 & 80$\pm$6 & 5$\pm$19 \\
100/20	& \cellcolor[gray]{0.8} 86$\pm$5 & 80$\pm$8 & 1$\pm$3 & 80$\pm$8 & 3$\pm$5 & 79$\pm$8 & 3$\pm$4 & 81$\pm$7 & 7$\pm$19 \\
250/20	& \cellcolor[gray]{0.8} 88$\pm$5 & 80$\pm$8 & 2$\pm$4 & 81$\pm$7 & 3$\pm$5 & 80$\pm$7 & 3$\pm$5 & 81$\pm$8 & 22$\pm$19 \\
500/20	& \cellcolor[gray]{0.8} 89$\pm$4 & 81$\pm$7 & 4$\pm$19 & 81$\pm$7 & 4$\pm$5 & 81$\pm$7 & 4$\pm$5 & 81$\pm$7 & 87$\pm$35 \\
\bottomrule
\end{tabular}
\end{center}
\label{tab:LBPvsRBC}
\end{table*}%


\subsection{Comparing Barcodes to other Methods} 
In the second series of experiments, we used \emph{leave-one-out validation} for different numbers of images $n_I$ to verify the performance of image-, barcode-, feature-, and hashing-based approaches when searching for the most similar image to build the consensus. Specifically, we used the structural similarity (SSIM) \cite{Zhou2004} to compare the query image $I_q$ with all other images in the atlas. (We also experimented with cross correlation, which was faster than SSIM but less accurate.) Further, the proposed atlas of barcodes was used to find the most similar barcode. Given the results from the previous sub-section, the RBC with incremental thresholding was selected as the best barcode approach to compare with other methods---insofar as it was slightly better than the RBC global. We also created an LBP histogram as a feature vector to find the most similar image via NNS. The Matlab code for this LBP implementation was taken from the Web \footnote{Coded by Marko Heikkil{\"a} and Timo Ahonen: http://www.cse.oulu.fi/CMV/Downloads/LBPMatlab}. Finally, we ran experiments for locality sensitive hashing (LSH). Specifically, we used the E$^2$LSH \cite{Andoni2006}\footnote{Matlab code by Greg Shakhnarovich, TTI-Chicago (2008): http://ttic.uchicago.edu/~gregory/download.html}. We resized the images to $32\times 32$ for E$^2$LSH, and we set the numbers of hash tables to 50 and the key size to 32 bits.

Any of these methods would then provide the user segments $S^1, S^2,\dots, S^{20}$ attached to the image they find. The consensus was then built, $\mathbf{C}=\textrm{STAPLE}(S^1,S^2,\dots,S^{20})$, and compared to the gold-standard image $G$ to calculate the accuracy of the computed consensus using the Jaccard index: 
 \begin{equation}
 J(\mathbf{C}, \mathbf{G})=\frac{|\mathbf{C} \cap \mathbf{G}|}{|\mathbf{C} \cup \mathbf{G}|}.
 \label{eq:jaccard}
 \end{equation} 
   
We also calculated the \emph{maximum achievable accuracy} (MAA). The results are presented in Table \ref{tab:ssimcorrfeature}. The average time $\bar{t}$ (in seconds) per query was also reported. The RBC was based on incremental thresholding. 

\textbf{Analysis of Table \ref{tab:ssimcorrfeature} --} We can observe from Table \ref{tab:ssimcorrfeature} that both accuracy and time increase as the size of the atlas grows (except with LSH). All methods reached a comparable level of accuracy with large atlases ($n_I=500$). The null hypothesis for accuracy measurements was rejected with $p=0.05$. As expected, image-based methods were the slowest. The RBC was the fastest method, despite being formally bounded by $\mathcal{O}(n_I)$. However, we are dealing with XOR operations, where a theoretical linear upper bound behaves like a sub-linear in practice. Many studies have shown that we can perform millions of Hamming distance calculations in less than a second \cite{Daugman1993,Liu2011}. As a hashing approach, LSH requires 15 times more time for search than the RBC when $n_I=500$.  If we increase the number of images in the atlas, we expect that LSH will outperform barcode-based methods at some point. The upper bound of LSH is, among others, a function of the number of collisions $N_c=\sum_{i=1:n_I}p^k (|| I_q-I_i ||)$, and this, in turn, is a function of the number of hash functions and tables (where $p$ denotes the probability of a collision of images) \cite{Andoni2006}. However, LSH requires increasing the number of hash tables and enlarging the key size, in terms of its bit length, in order to cope with larger data. We did run experiments for different numbers of hash tables, keeping the key size constant at 30 bits. With 50, 100, and 200 hash tables, we measured average times of $59\pm 7$ ms, $105\pm 10$ ms and $194\pm 10$ ms, respectively.  
  
\begin{table*}[htb]
\caption{Leave-one-out experiments for different image-retrieval methods for consensus building using $n_U=20$. Accuracy is measured by comparing the computed consensus $\mathbf{C}$ to the gold segment $\mathbf{G}$ using the Jaccard index $J(\mathbf{C},\mathbf{G})$. Maximum achievable accuracy (MAA) is provided for comparison. The average search times $\bar{t}$ are in seconds.}
\begin{center}
\begin{tabular}{@{}*{10}{c}@{}}
\toprule
 & & \multicolumn{2}{c}{Image-based} & \multicolumn{2}{c}{Barcode-based} & \multicolumn{2}{c}{Feature-based} & \multicolumn{2}{c}{Hashing-based}\\
 \cmidrule(r){3-4}\cmidrule(r){5-6}\cmidrule(r){7-8}\cmidrule(r){9-10}
$n_I/n_U$	&  \cellcolor[gray]{0.8} MAA & J$_\textrm{SSIM}$ & $\bar{t}_\textrm{SSIM}$ & J$_\textrm{RBC}$  & $\bar{t}_\textrm{RBC}$ & J$_\textrm{NNS}$  & $\bar{t}_\textrm{NNS}$ & J$_\textrm{LSH}$  & $\bar{t}_\textrm{LSH}$\\ 
\cmidrule(r){2-2}\cmidrule(r){3-4}\cmidrule(r){5-6} \cmidrule(r){7-8}\cmidrule(r){9-10}
10/20	& \cellcolor[gray]{0.8} 78$\pm$6 & 75$\pm$7 & 0.453 	& 76$\pm$6 & $<$0.001 	& 75$\pm$5 & 0.033 & -- & --\\
20/20	& \cellcolor[gray]{0.8} 80$\pm$6 & 78$\pm$7 & 0.879 	& 77$\pm$7 &$<$0.001 	& 80$\pm$7 & 0.021 & 79$\pm$ 7& 0.056 \\
50/20	& \cellcolor[gray]{0.8} 84$\pm$5 & 76$\pm$7 & 2.224 	& 77$\pm$7 & $<$0.001 	& 79$\pm$7 & 0.015 & 77$\pm$ 7 & 0.056 \\
100/20	& \cellcolor[gray]{0.8} 86$\pm$5 & 78$\pm$8 & 4.288 	& 80$\pm$8 & 0.003 	& 80$\pm$7 & 0.021 & 80$\pm$7 &  0.056\\
250/20	& \cellcolor[gray]{0.8} 88$\pm$5 & 80$\pm$8 & 10.902 	& 81$\pm$7 & 0.003 	& 81$\pm$8 & 0.045 & 80$\pm$7& 0.057\\
500/20	& \cellcolor[gray]{0.8} 89$\pm$4 & 81$\pm$8 & 21.534 	& 81$\pm$7 & 0.004 	& 81$\pm$8 & 0.087 & 81$\pm$7& 0.059\\
\bottomrule
\end{tabular}
\end{center}
\label{tab:ssimcorrfeature}
\end{table*}%

\textbf{The Accuracy of Computed Consensus --} 
We learned that the best we can do for an atlas of size 500 was MAA=$89\%\pm 4\%$ (Table \ref{tab:ssimcorrfeature}). The closest we came to this upper bound was with incremental RBC, achieving $81\%\pm 7\%$. These numbers represent the agreement between the computed consensus contours and the gold-standard segments that were available to us as a result of using synthetic images. However, it is not immediately clear what these numbers mean. To reinforce our findings, the accuracy of all 20 simulated users was measured against both the computed consensus $\mathbf{C}_q$ and the corresponding gold segment $\mathbf{G}_q$ (see Table \ref{tab:allusers}). Again, such a comparison would be impossible with real image data, insofar as there is no gold standard.

\begin{table}[htb]
\small
\caption{Leave-one-out validation with all images and gold segments, with the delineations of the 20 simulated users. The numbers report $\textrm{error}=| J(\mathbf{G},\mathbf{S}_i)-J(\mathbf{C},\mathbf{S}_i)|$ for segments $S_i$ of each user.}
\begin{center}
\begin{tabular}{@{}*{7}{c}@{}}
\toprule
 		& error  & & error & & error\\ 
 \cmidrule(r){2-2} \cmidrule(r){4-4}\cmidrule(r){6-6}
1	& $6\%\pm 4\%$ &  8	 & $9\%\pm 6\%$ &  15	& $8\%\pm	6\%$ \\
 2	& $7\%\pm 4\%$ &  9	 & $7\%\pm 5\%$ &  16	& $6\%	\pm 3\%$ \\
 3	& $8\%\pm 5\%$ &  10 & $9\%\pm 7\%$ &  17	& $9\%	\pm 6\%$ \\
 4	& $7\%\pm 6\%$ &  11	& $6\%	\pm 4\%$ &  18	& $8\%	\pm 5\%$ \\
 5	& $11\%\pm 7\%$ &  12	& $5\%\pm	4 \%$ &  19	& $10\%	\pm 6\%$ \\
 6	& $5\%\pm 5\%$ &  13	& $6\%\pm	4 \%$ &  20	& $10\%	\pm 7\%$ \\ 
 7	& $7\%\pm 5\%$ &  14	& $15\%\pm	6\%$ & &\\
\bottomrule
\end{tabular}
\end{center}
\label{tab:allusers}
\end{table}%

\textbf{Analysis of Table \ref{tab:allusers} --} The consensus contours $\mathbf{C}$ reached a high degree of overlap with the gold segments ($\approx 87\%\pm 6\%$). This demonstrates that the proposed registration-free barcode atlas can in fact deliver reliable results. Hence, $\mathbf{C}$ can evaluate the user segments with a total average error of $8\%\pm 5\%$ for all users. These numbers quantify how much a computational evaluation would differ were actual gold segments available.

\subsection{The Influence of Experience} 
It is pertinent to determine the influence that expertise has on the consensus. Of course, if we could assemble a large group of experts, then it would be likely that some of them would be highly qualified. Owing to practical constraints, however, we mostly rely on relatively small groups of users.   

We analyzed the performance of the 20 simulated users by comparing their segments to the gold segment (rather than to the consensus) to discover further insights into the overall results. Because the highest Jaccard was 89\%, the first group was slightly skewed to accommodate the best ``simulated'' users (Users 9, 11, and 19). For Users $1, 2,..., 20$ we defined 6 user groups, $U_A,\dots,U_F$ (see Table \ref{tab:usergroupsdefined}). 

\begin{table}[htb]
\small
\caption{Grouping the simulated users based on their contouring expertise, measured by comparing their contours $S_i^j$ with the gold segment $G_i$ for $i\in\{1, 2,\dots, 500\}$ and $j\in\{1, 2,\dots, 20\}$.}
\begin{center}
\begin{tabular}{@{}*{2}{l}@{}}
\toprule
$J(G,S_i)$ & User Group \\ 
\cmidrule(r){1-1}\cmidrule(r){2-2}
$[85\%,100\%]$ & $U_A	= \{9,11,19\}$\\ 
$[80\%,85\%)$ & $U_B 	= \{3,4,8,14\}$ \\
$[70\%,80\%)$ & $U_C 	= \{1,2,5,6,20\}$ \\
$[60\%,70\%)$ & $U_D 	= \{7,12\}$ \\
$[50\%,60\%)$ & $U_E 	= \{10,15,16,17,18\}$ \\
$[0\%,50\%)$ & $U_F 	= \{13\}$ \\ 
\bottomrule
\end{tabular}
\end{center}
\label{tab:usergroupsdefined}
\end{table}%

The experiments described above demonstrated that when previously segmented images from a large number of users are available, the barcode atlas does deliver a reasonable approximation of the user assessment via the computed consensus. To identify the effect of user expertise, we ran experiments in cases where only a small number of users were available (see Table \ref{tab:GOODBAD}). These results indicate that, when only a few users (i.e., their atlases) are available, the quality of the computed consensus depends heavily on their expertise.   

\begin{table}[htb]
\small
\caption{Effect of user expertise on the computed consensus: three random trials with different numbers of images. }
\begin{center}
\begin{tabular}{@{}*{6}{c}@{}}
\toprule
\multicolumn{2}{c}{50 images} & \multicolumn{2}{c}{100 images} & \multicolumn{2}{c}{250 images} \\
\cmidrule(r){1-2}\cmidrule(r){3-4}\cmidrule(r){5-6}
Users & $J$ & Users & $J$ & Users & $J$ \\
\cmidrule(r){1-1}\cmidrule(r){2-2}\cmidrule(r){3-3}\cmidrule(r){4-4}\cmidrule(r){5-5}\cmidrule(r){6-6}
$[BBC]$ 	& $91\pm 3$ & $[ABD]$ 	& $87\pm 5$ & $[AAB]$ 	& $88\pm 4$ \\
$[CCE]$	& $71\pm 4$ & $[ADE]$ 	& $69\pm 7$ & $[AEF]$ 	& $55\pm 5$ \\
$[AAF]$	& $82\pm 6$ & $[ECC]$ 	& $61\pm 4$ & $[EEF]$ 	& $52\pm 5$ \\
\bottomrule
\end{tabular}
\end{center}
\label{tab:GOODBAD}
\end{table}%

\subsection{Testing with Real Images}
\label{sec:realimages}
As mentioned above, validating the consensus approach poses an inherent challenge: if the consensus serves as a gold standard, then how can we measure its own accuracy? 

In the previous section, we simulated images in order to exploit ``perfect segments''.  In this section, we validate our approach using actual MR images of prostates. The image data from 15 patients were manually delineated by 5 oncologists. 

The MR images used in this study were derived from an online database\footnote{\url{http://prostatemrimagedatabase.com/}}. The database contains T2-weighted MR volume datasets, provided by Brigham and Women's Hospital, the National Center for Image-guided Therapy, and Harvard Medical School. The images comprised T2-weighted MR images (T2W-MR) with endorectal coils. The pulse-sequence groups in the DICOM headers of most of the T2-weighted images were marked fast-spin echo (FSE), although some were marked as fast-relaxation fast-spin echo-accelerated (FRFSE-XL). The dataset contained images with slice thickness ranging from 2.5mm to 4.0mm, and varying contrast levels and signal-to-noise characteristics. All of the images were captured at a depth of 16 bits, and they varied in size from 256$\times$256 to 512$\times$512 pixels. 

\begin{figure*}[htb]
\begin{center}
\includegraphics[width=2.1in,height=2.1in]{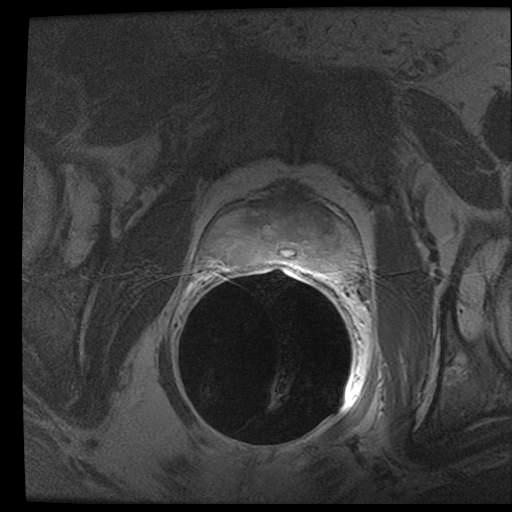}
\includegraphics[width=2.1in,height=2.1in]{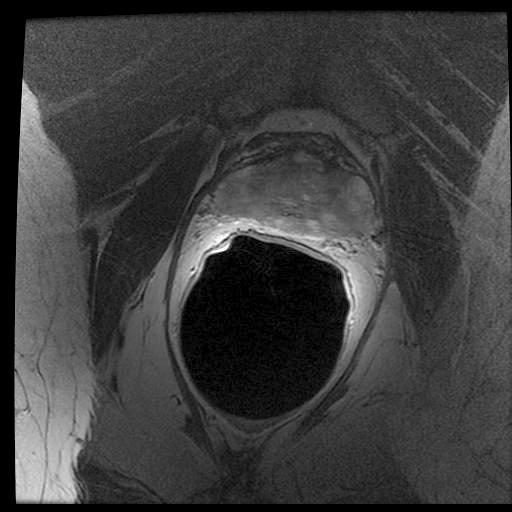} 
\includegraphics[width=2.1in,height=2.1in]{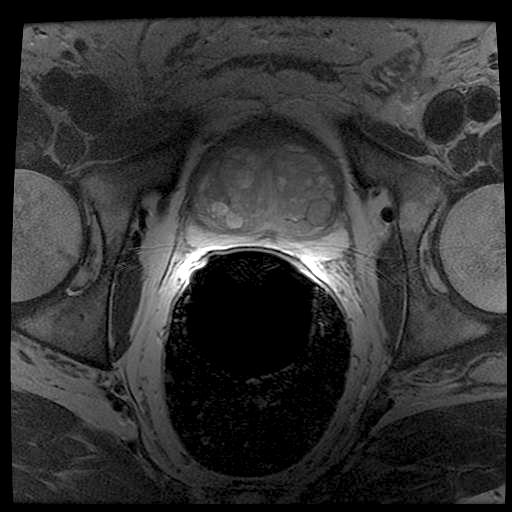} \\ \vspace{0.05in}
\includegraphics[width=2.1in,height=2.1in]{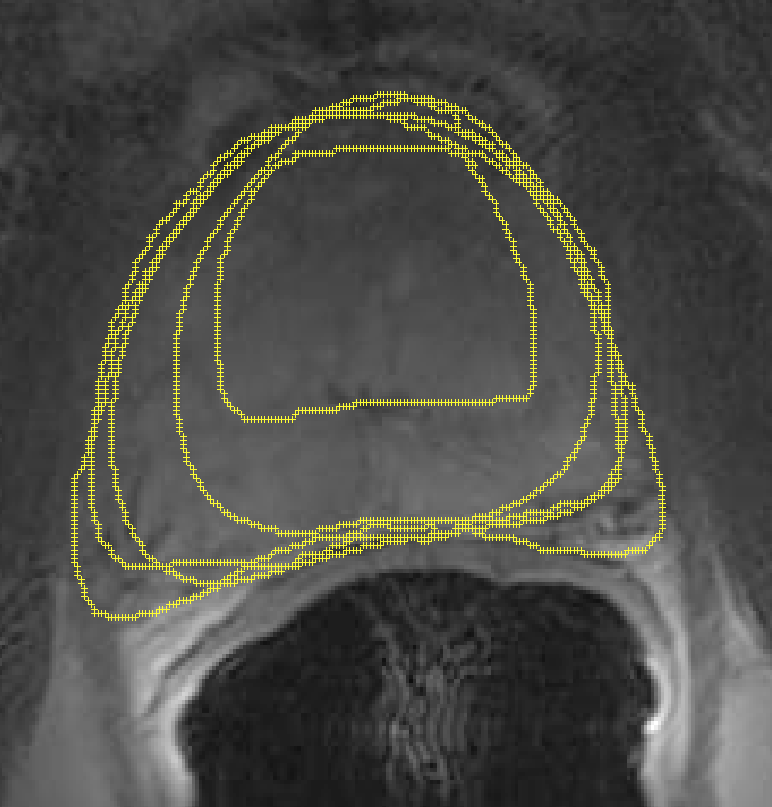}
\includegraphics[width=2.1in,height=2.1in]{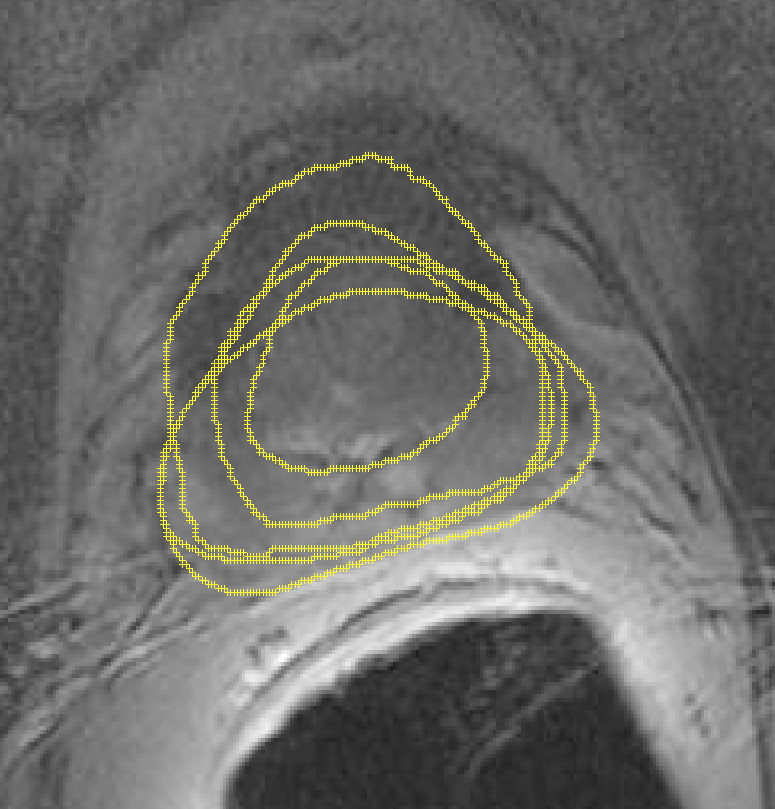}
\includegraphics[width=2.1in,height=2.1in]{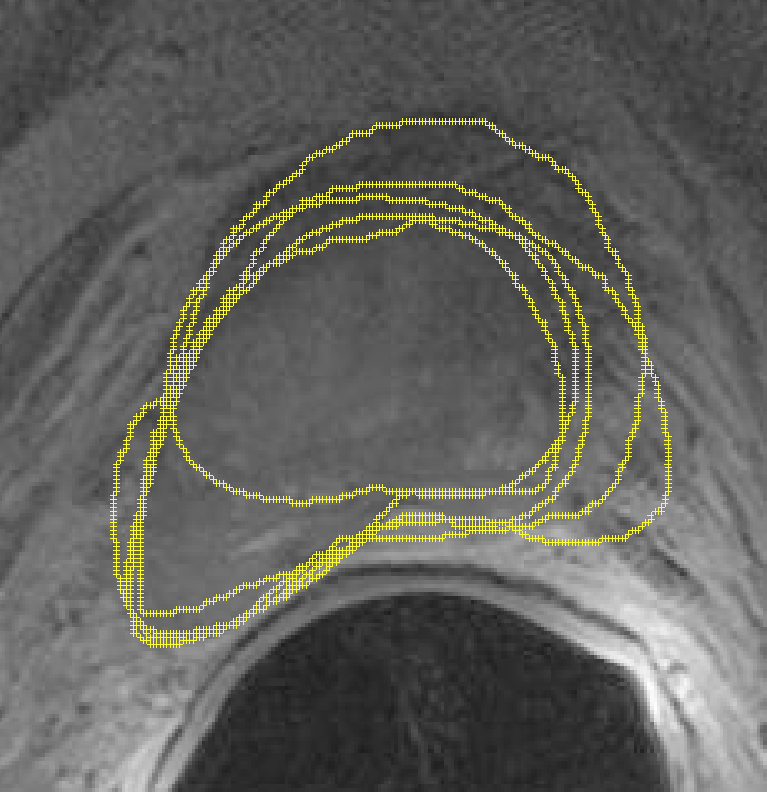}
\caption{Top: Sample MRI slices from different patients. Bottom: considerable variability among  5 oncologists.}
\label{fig:MRIsample}
\end{center}
\end{figure*}

Sample images are depicted in Figure \ref{fig:MRIsample} (top row). Generally, one assumes that prostate segmentation is a relatively easy task. However, the variability of such segmentation remains considerable, and this is conspicuous in Figure \ref{fig:MRIsample} (bottom row). 

We randomly selected 15 patients (out of more than 100) with a total of 558 slices, from which 145 slices were contoured by all 5 oncologists, resulting in a total of 725 segments\footnote{All DICOM images and their manual segmentations were provided by Segasist Technologies, Waterloo, ON, Canada.}. Similar to the validation using simulated images, we first ran STAPLE on all user segments to generate a consensus for each slice. (Note that this is a ``regular'' consensus, insofar as all experts were available to mark the same image.) After we have the \emph{regular} consensus, we can measure the agreement of each user using this consensus. This basically measures the extent to which each user has contributed to the consensus for that image.  Table \ref{tab:con5Users} reports the results. From these results, User 3 is the most accurate\footnote{Theoretically, it is possible for four bad users to dominate the consensus, such that the fifth (excellent) user is understood as the worst. This is another reason to favor a large number of experts when building a consensus.}. We selected User 3 as the gold standard. Then, we eliminated the segments from User 3 when building the atlas, in order to measure the accuracy of the computed consensus contours by comparing it against the manual delineations from User 3.     

\begin{table}[htb]
\small
\caption{Contouring accuracy of individual oncologists, measured against the consensus contour.}
\begin{center}
\begin{tabular}{@{}*{7}{c}@{}}
\toprule
	& U1	& U2	& U3	& U4	& U5	& Best  \\
 \cmidrule(r){2-2} \cmidrule(r){3-3}\cmidrule(r){4-4}\cmidrule(r){5-5}\cmidrule(r){6-6}\cmidrule(r){7-7}
Patient 1	& 81	& 61	& 94	& 82	& 82	& U3 \\
Patient 2	& 90	& 42	& 90	& 17	& 87	& U1,U3 \\
Patient 3	& 66	& 58	& 99	& 75	& 67	& U3 \\
Patient 4	& 93	& 60	& 82	& 72	& 87	& U1 \\
Patient 5	& 94	& 82	& 89	& 79	& 83	& U1 \\
Patient 6	& 90	& 71	& 65	& 78	& 65	& U1 \\
Patient 7	& 85	& 90	& 87	& 81	& 87	& U3,U5 \\
Patient 8	& 88	& 64	& 94	& 75	& 86	& U3 \\
Patient 9	& 90	& 81	& 99	& 87	& 89	& U3 \\
Patient 10	& 96	& 83	& 95	& 88	& 88	& U1 \\
Patient 11	& 88	& 70	& 96	& 80	& 75	& U3 \\
Patient 12	& 95	& 69	& 95	& 90	& 90	& U1,U3 \\
Patient 13	& 97	& 43	& 95	& 72	& 85	& U1 \\
Patient 14	& 85	& 29	& 96	& 71	& 69	& U3 \\
Patient 15	& 94	& 23	& 95	& 73	& 90	& U3 \\
\bottomrule
$\mu (\%)$		& 89	& 62	& 91	& 75	& 82	&  \\
$\sigma (\%)$ & 8 & 20 & 9 & 17 & 9 & \\
\bottomrule
\end{tabular}
\label{tab:con5Users}
\end{center}
\end{table}

We used the RBC with incremental thresholding, because this method emerged as the best after our previous experiments with simulated images. Upon building the atlas of barcodes using segments from Users 1, 2, 4, and 5, (excluding User 3), we repeated the experiments with real images to measure the accuracy of the computed consensus contours when the segments of User 3 were employed as gold-standard segments. The results are reported in Table \ref{tab:comCon4Users}. A Jaccard accuracy of $\approx\!87\!\pm\!9$ for 15 patients shows that the computed consensus is viable. To once again determine the actual quality of the consensus, Table \ref{tab:UserModellignErrorReal} reports the user-modelling errors. 

\begin{table}[htb]
\small
\caption{Accuracy of the computed consensus for MR images.}
\begin{center}
\begin{tabular}{@{}*{4}{c}@{}}
\toprule
	& $n_\textrm{I}$	& $J$	& $\bar{t}$(ms)	 \\
 \cmidrule(r){2-2} \cmidrule(r){3-3}\cmidrule(r){4-4}
 5 Patients	& 50 		& 84.4\%$\pm$11\% 	& 1$\pm$3 \\
10 Patients	& 94		& 85.7\%$\pm$10\% & 1$\pm$3 \\
15 Patients	& 145	& 86.7\%$\pm$9\% 	& 1$\pm$4 \\
\bottomrule
\end{tabular}
\label{tab:comCon4Users}
\end{center}
\end{table}

\begin{table}[htb]
\small
\caption{$\textrm{error}=| J(\mathbf{G},\mathbf{S}_i)-J(\mathbf{C},\mathbf{S}_i)|$ }
\begin{center}
\begin{tabular}{@{}*{2}{c}@{}}
\toprule
User 1	& 7.5\%$\pm$6.5\% \\
User 2	& 6.5\%$\pm$6.9\% \\
\textbf{User 3}	& [\textbf{\emph{gold standard}}] \\
User 4	& 9.0\%$\pm$7.6\% \\
User 5	& 3.4\%$\pm$5.6\% \\
\bottomrule
\end{tabular}
\label{tab:UserModellignErrorReal}
\end{center}
\end{table}

\textbf{Barcode of images versus barcode of ROIs --} The only difference between the experiments that used synthetic images and those that used real MR images was the manner by which the barcodes were calculated. In the case of synthetic images, most of the image area was relevant to encoding the prostate gland. In the MR images, by contrast, much more information is depicted, and the barcode loses its expressiveness when calculating the entire image. We assumed that when a user requests a consensus contour (as a second opinion or ``peer review'', as it were) that user must have either already delineated the prostate, or at least drawn a rectangle around the ROI. That way, we can easily extract the barcode for the prostate region. Furthermore, we enlarged the dimensions of the bounding box constructed around each user's segment by $30$ pixels ($\approx 3$cm) to capture some of the structures around the prostate gland. 

\section{Discussion}
The main idea and algorithm proposed in this work to deliver consensus contours via the search for barcodes is apparently promising as the results from the previous section demonstrate. However, the computational consensus has some requirements and limitations:
\begin{itemize}
\item The present approach only deals with 2D segmentation. Although many applications, e.g., radiation planning, are widely focused on segmentation of individual slices, an extension to 3D volumes would be necessary at some point.
\item Building consensus contour with a barcode-based atlas search only works when \emph{big image data} is available. That means in practice we need a very large number of studies (images of different patients) that offer the possibility of finding really similar cases. 
\item One might be able to work with not-very-large datasets as well. However, this would require that every study (images of the same patient) should be delineated by multiple experts such that locating similar cases is reduced to consensus building among multiple experts who have actually marked the same images.       
\end{itemize}
     
\section{Conclusions}
In this paper, we introduced the idea of using ``barcodes'' to facilitate consensus building. The Radon barcodes capture the content of an image, hence, providing a novel approach to annotate digital images. Through fast XOR operations to find similar barcodes, we demonstrated that similar cases can be retrieved in order to compute consensus contours when the atlas contains manual segmentations from multiple users (or in cases where there are multiple atlases). To validate the proposed technique, we used synthetic images, from which perfect segmentations can be used to quantify the accuracy of the computed consensus. Moreover, we used T2-weighted prostate MR images of 15 patients with markings from 5 oncologists to run additional experiments. The results appear to be promising. 

We worked exclusively with the first hit, and we did not use any registration. It will be pertinent to investigate the effect of registration in future research, when multiple similar cases are used simultaneously to compute the consensus. Combining the barcode atlas with other works, such as locality-sensitive hashing, is also worth investigating.  

Different approaches to encode projections (i.e., binarization of projects) has to be investigated to minimize the loss of information. Methods like recently proposed \emph{MinMax} Radon barcodes \cite{tizhoosh2016minmax} may produce better results because they preserve the changes in the projection shape.  


\section{Availability of data and material}
The image data used in this paper (both synthetic and real) will be available for download under \url{http://kimia.uwaterloo.ca}. 

\section{Acknowledgements}
We would like to thank Dr. Simon Warfield, Dr. Issam El Naqa, and Dr. Aditya Apte for answering our questions regarding STAPLE and its CERR implementation. We would also like to thank Segasist Technologies (Waterloo, Canada) for providing MR images and their segments.  This project was supported by an NSERC Discovery grant.


\bibliographystyle{plain}
\bibliography{egbib}
\end{document}